%% file: main.tex
\documentclass[letterpaper,twocolumn,fleqn]{article} 

\usepackage{ist}
\usepackage{amsmath,graphicx}
\usepackage{float}

\usepackage{subcaption}

\usepackage{mwe}
\usepackage{multirow}
\usepackage{color}
\usepackage{xcolor}
\usepackage{hyperref}

\usepackage{pgfplots}
\usepgfplotslibrary{fillbetween}
\usepackage{filecontents}

\usepackage{nicefrac}
\usepackage{float}
\usepackage[normalem]{ulem}
\usepackage{lipsum}
\usepackage{booktabs}   % for top/mid/bottom rules
\usepackage{cleveref}
\usepackage{arydshln}
\definecolor{best}{HTML}{ffb2b2}
\definecolor{secondbest}{HTML}{ffd9b2}
\definecolor{thirdbest}{HTML}{ffffb2}

\title{A General Pipeline for Dense Illuminant Estimation via Physically Based Synthetic Data}

\author{ Luca Cogo, Gianmarco Corti, Simone Bianco, Raimondo Schettini\\Department of Informatics, Systems and Communication, University of Milano -- Bicocca, Milan 20126, Italy}

\date{} % date has an empty field.

\begin{document} 

\maketitle 

\thispagestyle{empty} % prevents the first page to be numbered

%%%%%%%%%%%%%%%%%%%%%%%%%%%%%%%%%%
% Abstract
%%%%%%%%%%%%%%%%%%%%%%%%%%%%%%%%%%

\input{sections/01_abstract}

\input{sections/02A_introduction}
\input{sections/02B_related_work}

\input{sections/03_method}
\input{sections/04_experimental_results}

\input{sections/05_conclusion}

%%%%%%%%%%%%%%%%%%%%%%%%%%%%%%%%%%
% Bibliography
%%%%%%%%%%%%%%%%%%%%%%%%%%%%%%%%%%
\small{
\bibliographystyle{IEEEbib}
\bibliography{refs}
}

\end{document}

%% file: sections/01_abstract.tex
\begin{abstract}
Illuminant estimation is a fundamental problem in computational photography, as it enables the correction of color shifts induced by varying lighting conditions. While learning-based methods have demonstrated strong performance, their progress is hindered by the limited availability of large-scale datasets with accurate illuminant ground-truth. In this work, we propose a general and reusable pipeline to derive dense illuminant chromaticity maps from physically based 3D-rendered scenes. By repurposing an existing 3D scene collection, our approach enables the systematic generation of pixel-wise illuminant annotations under controlled lighting conditions, effectively lowering the barrier to data acquisition for learning-based illuminant estimation. Using this pipeline, we generate a large-scale synthetic set of 74,321 images, which we employ for pre-training both single- and multi-illuminant estimation models. 
Extensive experiments with state-of-the-art architectures show that synthetic pre-training consistently improves performance, with gains of up to 28\% for single-illuminant estimation and up to 57\% for multi-illuminant estimation, particularly in data-scarce regimes. These findings demonstrate that synthetic data generation pipelines offer an effective and scalable solution for the pre-training of illuminant estimation methods.
% Extensive experiments with state-of-the-art architectures demonstrate that synthetic pre-training consistently improves performance, yielding gains of up to 28\% for single-illuminant estimation and up to 57\% for multi-illuminant estimation. Improvements are particularly pronounced in data-scarce regimes.These results highlight the effectiveness of synthetic data generation pipelines as a scalable strategy to alleviate the limitations imposed by scarce annotated data.
\end{abstract}

%% file: sections/02A_introduction.tex
\section{Introduction}
\label{sec:intro}

Illuminant estimation is a key component of modern digital imaging pipelines, as it compensates for color shifts caused by varying illumination conditions. By estimating the chromaticity of the scene’s light sources, illuminant estimation enables color correction that aligns image appearance with human perception of white and neutral tones under different lighting. Beyond photographic enhancement, accurate illuminant estimation is essential for many computer vision tasks, including object recognition, scene understanding, and image retrieval, where color consistency directly affects downstream performance~\cite{ebner2020color}.
Early approaches relied on global statistical assumptions about the color distribution of natural images~\cite{van2005edge}. While computationally efficient, such methods often fail under complex or spatially varying illumination. Learning-based models, including convolutional neural networks~\cite{bianco2015color, hu2017fc4, barron2017fast} and more recent transformer-based architectures~\cite{kim2024attentive}, significantly improve robustness by leveraging contextual and semantic information. However, their effectiveness is constrained by the limited availability of large-scale datasets with accurate illuminant ground-truth, whose acquisition requires controlled capture conditions and reference measurements.
This limitation is especially pronounced for multi-illuminant estimation, where illumination varies spatially across the scene and pixel-level annotations are required.
To address this challenge, we leverage physically based 3D-rendered scenes as a principled source of dense illuminant supervision. Photorealistic renderings provide explicit access to reflectance and illumination properties of the scenes, enabling the derivation of accurate per-pixel illuminant chromaticity, information that is prohibitively difficult to obtain at scale from real images.
In this work, we introduce a general and reusable methodology for extracting dense illuminant chromaticity maps from physically based 3D renderings. The proposed pipeline exploits the illumination–reflectance decomposition available in modern rendering engines to generate physically grounded per-pixel supervision, suitable for pre-training both single- and multi-illuminant estimation models and independent of network architecture.
As a concrete instantiation of the proposed methodology, we apply the pipeline to the Hypersim dataset~\cite{roberts2021hypersim} to generate a large synthetic corpus for pre-training. Importantly, we emphasize that this instantiation serves as a proof of scalability rather than a dataset-specific contribution: the methodology itself is readily transferable to other rendering environments and synthetic scene collections. Extensive experiments with state-of-the-art architectures~\cite{ronneberger2015u, gharbi2017deep, hu2017fc4} and standard benchmark datasets~\cite{kim2021large, beigpour2013multi, hemrit2018rehabilitating} demonstrate consistent performance improvements across architectures and illuminant estimation settings.

%% file: sections/02B_related_work.tex
% \begin{figure*}[ht]
%     \centering
%     % Immagine più grande a sinistra
%     \begin{minipage}[r]{0.23\textwidth}
%         \centering
%         \begin{subfigure}{\textwidth}
%             \includegraphics[width=1.0\linewidth]{images/images_mostra_4.png}
%             \caption{Input}
%             \label{fig:images_mostra_4}
%         \end{subfigure}
%     \end{minipage}
%     %\hspace{0.1cm}
%     % Immagini più piccole a destra
%     \begin{minipage}[l]{0.23\textwidth}
%         \centering
%         \begin{subfigure}{\textwidth}
%             \includegraphics[width=1.0\linewidth]{images/chroma_mostra_4.png}
%             \caption{Illuminant map}
%             \label{fig:chroma_mostra_4}
%         \end{subfigure}
%     \end{minipage}
%     \caption{Sample of images from the dataset. Figure (a) contains the RGB images used as input of the model. Figure (b) shows the respective illuminant images used as ground-truth.}
%     \label{fig:mostra_4}
% \end{figure*}

\section{Related Work}
\label{sec:related_work}
\subsection{Illuminant Estimation Methods}
% \begin{figure*}
%     \centering
%     \includegraphics[width=\linewidth]{assets/example_hypersim.pdf}
%     \caption{Sample images from the dataset. (a) illustrates the individual image components. (b) presents the input images along with their corresponding ground-truth illuminant images.}
%     \label{fig:mostra_4}
% \end{figure*}
\begin{figure*}
    \begin{subfigure}{\linewidth}
    \centering
    \includegraphics[width=0.7\linewidth]{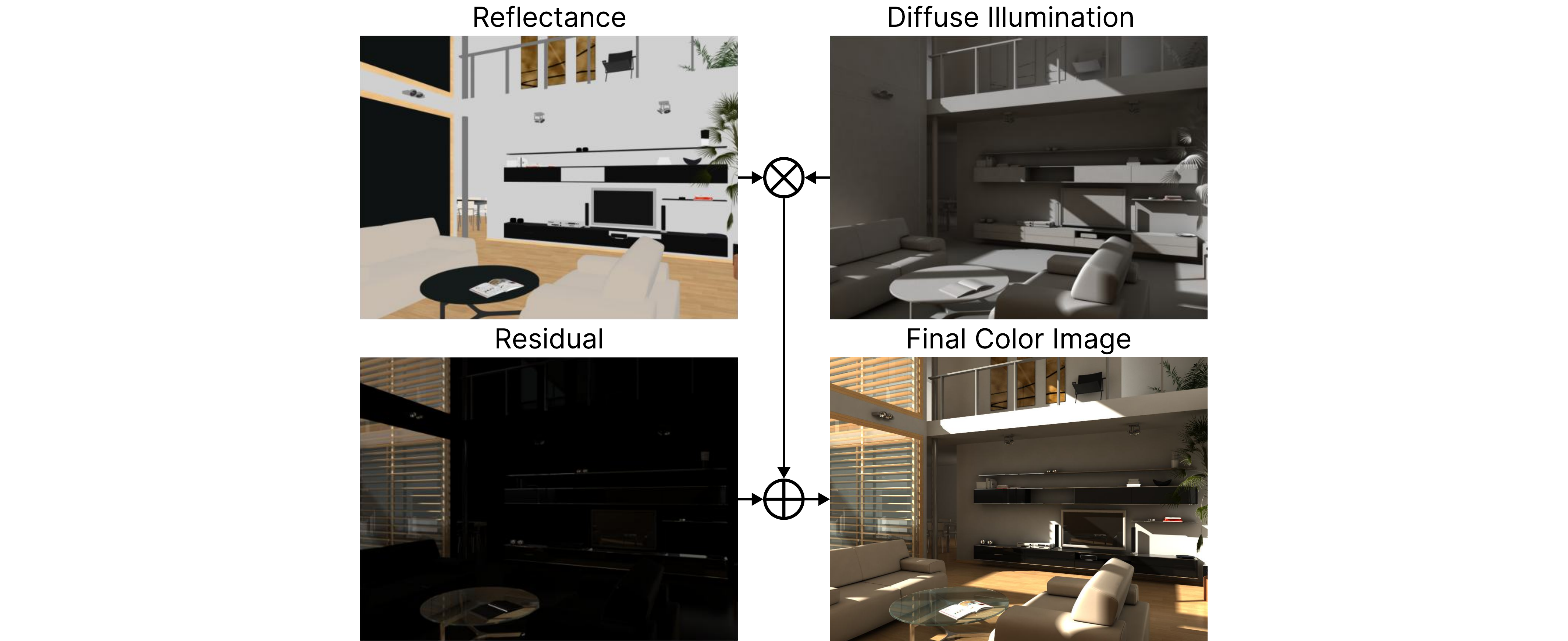} 
    \subcaption{\vspace{4mm}}
    \end{subfigure}
    \begin{subfigure}{\linewidth}
    \centering
    \includegraphics[width=0.7\linewidth]{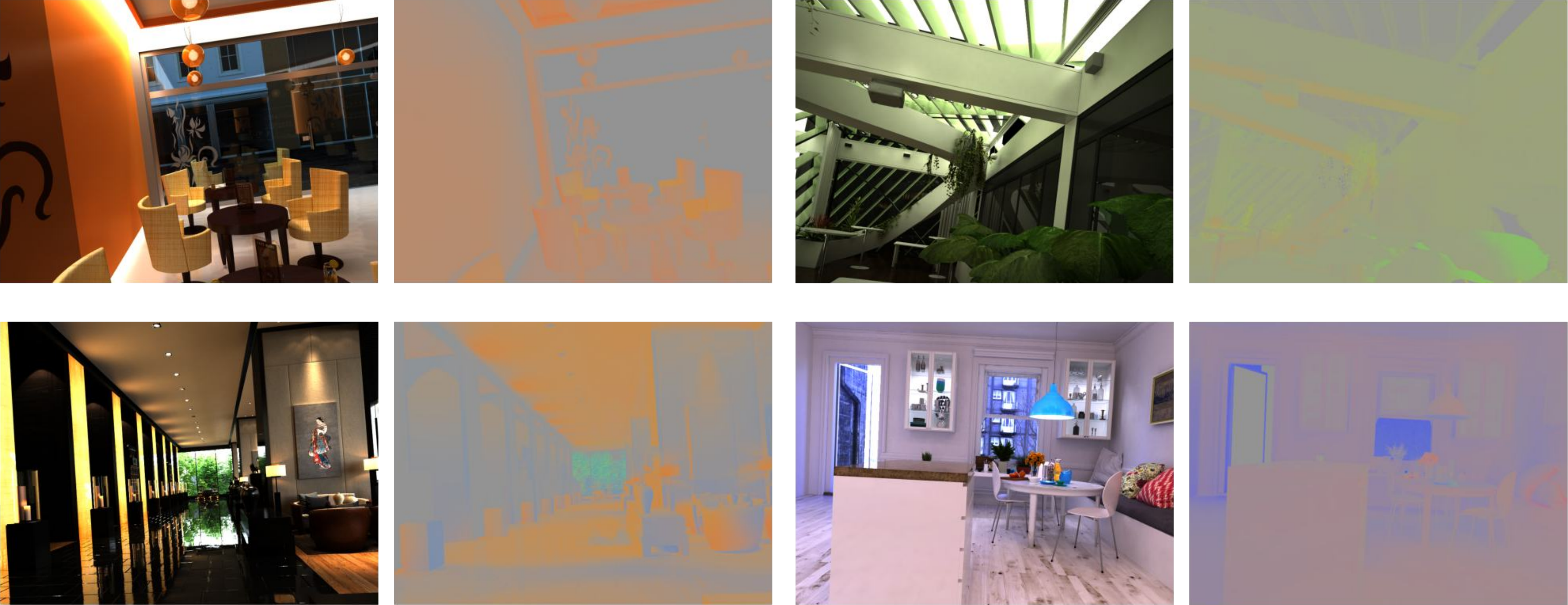} 
    \subcaption{\vspace{4mm}}
    \end{subfigure}
    \caption{Sample images from the dataset. (a) illustrates the individual image components. (b) presents the input images along with their corresponding ground-truth illuminant images.}
    \label{fig:mostra_4}
\end{figure*}

\label{sec:awb_methods}

Existing approaches for illuminant estimation can be broadly categorized into traditional statistics-based methods and learning-based techniques, and further distinguished by whether they assume a single global illuminant or spatially varying (multi-illuminant) illumination. Early methods typically assume a single dominant illuminant affecting the entire scene. Classical algorithms such as Gray World, White Patch, Shades of Gray, and Grey Edge~\cite{van2005edge} rely on global statistical assumptions about the color distribution of natural images. While computationally efficient, these approaches often fail in complex scenes with spatially non-uniform lighting or multiple light sources.
Learning-based methods have significantly advanced illuminant estimation by leveraging data-driven representations. CNN-based models, such as those proposed by Bianco et al.~\cite{bianco2015color} and the FC$^4$ model by Hu et al.~\cite{hu2017fc4}, directly regress illuminant chromaticity from image content and substantially outperform traditional techniques in single-illuminant settings. To handle more realistic scenarios with spatially varying illumination, recent work has focused on dense or local estimation. Encoder–decoder architectures, including UNet~\cite{ronneberger2015u}, predict pixel-wise or region-wise illuminant maps~\cite{kim2021large}, while models such as HDRNet~\cite{gharbi2017deep} apply locally adaptive color transformations guided by learned illumination representations. More recently, transformer-based architectures~\cite{kim2024attentive} have further improved performance by capturing long-range spatial dependencies and contextual cues. Despite these advances, learning-based methods remain heavily dependent on large, accurately annotated datasets, which are challenging to acquire in real-world settings.

\subsection{Illuminant Estimation Datasets}
\label{sec:awb_datasets}

Research in illuminant estimation has historically relied on a limited number of real-world datasets. Widely used benchmarks such as ColorChecker~\cite{hemrit2018rehabilitating}, NUS-8~\cite{cheng2014illuminant}, and Cube++~\cite{ershov2020cube++} provide images captured under controlled conditions, with illuminant annotations obtained from calibration targets. These datasets are primarily designed for single-illuminant estimation and typically provide only global illuminant annotations.
To better capture real-world lighting complexity, newer datasets such as LSMI~\cite{kim2021large} and MIMO~\cite{beigpour2013multi} introduce scenes with multiple light sources and offer pixel-wise illuminant ground-truth. While these datasets enable the development of multi-illuminant estimation methods, their scale and diversity remain limited due to the cost and complexity of acquiring accurate annotations. Dense ground-truth illuminant maps often require specialized acquisition setups, careful calibration, or manual intervention, resulting in datasets that are relatively small and heterogeneous in annotation quality. Consequently, the scarcity of large-scale, diverse, and consistently annotated real-world datasets remains a major bottleneck for training and evaluating modern illuminant estimation models.

\subsection{Synthetic Datasets for Vision Tasks}
\label{sec:synthetic_dataset_vision_task}

Synthetic datasets have become an important resource for many computer vision tasks, including optical flow~\cite{mayer2016large}, semantic segmentation~\cite{richter2016playing}, and depth estimation~\cite{ros2016synthia}. By providing full control over scene geometry, material properties, and illumination, synthetic data enable the creation of large-scale datasets with precise ground-truth annotations at minimal cost. Prior work has shown that pre-training on synthetic data followed by fine-tuning on real-world datasets can significantly improve generalization and robustness~\cite{ros2016synthia, gaidon2016virtual}.
Advances in photorealistic rendering and physically based illumination models have further reduced the gap between synthetic and real imagery, allowing realistic simulation of complex lighting interactions and material reflectance. Despite these successes, the use of synthetic data for illuminant estimation remains limited. Early attempts relied on simplified illumination models or global color transformations~\cite{gijsenij2011computational}, which fail to capture the intricate interactions between lighting, geometry, and surface reflectance.
More recently, photorealistic 3D datasets such as Hypersim~\cite{roberts2021hypersim} and Replica~\cite{straub2019replica} provide detailed geometric, material, and illumination information that can be repurposed for illuminant estimation. These datasets offer a promising foundation for pre-training and benchmarking illuminant estimation models, particularly in scenarios where annotated real-world data are scarce.

%% file: sections/03_method.tex
\input{tabs/quantitative_multi}

\section{Data Generation Pipeline}
\label{sec:proposed_dataset}

Hypersim~\cite{roberts2021hypersim} is a synthetic HDR dataset comprising 77,400 photorealistic indoor images rendered from 461 distinct 3D scenes. Originally developed for geometric learning, physically based vision tasks, and multi-task scene understanding, Hypersim provides detailed geometric and photometric metadata for each scene. This rich information makes the dataset suitable for tasks beyond its original scope, including illuminant estimation.
Each rendered image in Hypersim is accompanied by multiple data modalities, including geometric attributes and a decomposition of the scene radiance into diffuse illumination and residual non-diffuse components. The diffuse illumination images are particularly relevant for illuminant estimation, as they represent scene lighting independent of surface reflectance and can serve as a reference for estimating illuminant chromaticity and constructing dense illumination maps.

Although Hypersim does not natively provide illuminant annotations for color constancy or illuminant estimation, its physically based rendering ensures photometric consistency between illumination, reflectance, and observed color. This allows us to derive dense ground-truth illuminant maps from the available rendering outputs. Specifically, each image is defined by three components: diffuse illumination ($L$), diffuse reflectance ($R$), and a residual term ($\rho$) capturing specular reflections, interreflections, and other non-Lambertian effects. The final rendered image ($I$) is reconstructed as:
\begin{equation}
I = (R \odot L) + \rho.
\label{eq:color_image_equation}
\end{equation}
In this work, we use the diffuse illumination component $L$ as a proxy for the scene illuminant. In principle, $L$ provides an accurate estimate of the illuminant for Lambertian surfaces. However, in practice, relying on $L$ alone is insufficient: in regions dominated by specular reflections, shadows, or other non-diffuse effects, the residual term $\rho$ often contributes most of the observed radiance, while $L$ may contain near-zero or missing values (rendered as black pixels). Direct use of $L$ would therefore produce incomplete or physically inconsistent illuminant maps.
To address this, we augment the diffuse illumination with the grayscale intensity of the residual term prior to normalization:
\begin{equation}
\hat{L} = \text{normalize}\bigl(L + \text{avg}(\rho)\bigr),
\end{equation}
where $\text{avg}(\rho)$ denotes the per-pixel mean over the RGB channels of the residual image. The residual is assumed to be dominated by view-dependent reflectance and therefore provides limited reliable chromatic information. This choice is intentionally conservative with respect to chromaticity, as it avoids injecting spurious color information while preserving illumination magnitude in non-diffuse regions. The combined illumination is then normalized to obtain dense per-pixel chromaticity maps.
All images are provided in HDR format. We apply the tonemapping function released by the dataset authors, omitting the gamma correction step to preserve a linear relationship between radiance and pixel values. In cases where the diffuse illumination image contains black pixels—indicating regions where illumination is fully captured by the residual term—we assume neutral illumination and generate binary masks to exclude these pixels from the training loss.
A manual inspection is carried out to remove malformed images, resulting in a final set of 74,321 images.
\Cref{fig:mostra_4} shows examples of reconstructed scenes alongside their corresponding dense illuminant chromaticity maps. 
The resulting dataset, which we refer to as Hypersim-WB, is split following a per-scene strategy to avoid data leakage: 80\% of the scenes are used for training, 20\% for testing, with an additional 20\% of the training set reserved for validation.

%% file: tabs/quantitative_multi.tex
\begin{table*}[]
\caption{Quantitative results, in terms of angular error, of the considered methods on three multi-illuminant estimation datasets: the proposed Hypersim-WB, LSMI-Galaxy~\cite{kim2021large}, and MIMO~\cite{beigpour2013multi}. For each learning-based method we compare a version trained from-scratch and one that is fine-tuned after pre-training on the Hypersim-WB dataset. We highlight in bold the per-method best and underline the per-dataset best.\vspace{2mm}}
\label{tab:main_results}
\resizebox{\linewidth}{!}{
{\sffamily
\begin{tabular}{lccccccccccccc}
\toprule
                                            &                                           & \multicolumn{4}{c}{Hypersim-WB}   & \multicolumn{4}{c}{LSMI-Galaxy~\cite{kim2021large}}  & \multicolumn{4}{c}{MIMO~\cite{beigpour2013multi}}                 \\ \cmidrule(lr){3-6} \cmidrule(lr){7-10} \cmidrule(lr){11-14} \\\\\noalign{\vskip -2.25em}
Method                                       & Pre-training Data    & Mean                  & B-25             & W-25                & 95-P              & Mean                  & B-25             & W-25                & 95-P          & Mean                  & B-25             & W-25                & 95-P               \\ \midrule
GW~\cite{van2005edge} & -                   & 5.45   & 1.88   & 11.10  & 13.2  & 4.18                   & 0.95                   & 8.47                   & 10.30                   & 2.33                   & 0.89                   & 4.20                    & 4.83                   \\\ \\\noalign{\vskip -1em} \cdashline{1-14} \\\noalign{\vskip -1em}
\multirow{2}{*}{UNet~\cite{kim2021large}}                        & -                    & \uline{3.18}   & \uline{1.1}    & \uline{6.35}  & \uline{8.00}     & 1.94                   & 0.65                   & 3.88                   & 4.63                   & 3.30                    & 2.06                    & 5.48                   & 6.08                   \\
                                             & Hypersim-WB          & -      & -      & -     & -     & \textbf{\uline{1.83}} & \textbf{0.61}          & \textbf{\uline{3.65}} & \textbf{\uline{4.06}} & \textbf{\uline{2.13}} & \textbf{\uline{0.83}} & \textbf{\uline{4.06}} & \textbf{\uline{4.71}} \\\ \\\noalign{\vskip -1em} \cdashline{1-14} \\\noalign{\vskip -1em}
\multirow{2}{*}{HDRNet~\cite{gharbi2017deep}}                      & -                    & 3.77   & 1.45   & 7.57  & 9.20   & 2.36                   & 0.67                   & 4.98                   & 5.75                   & 3.41                   & 2.31                    & \textbf{4.87}          & \textbf{5.34}          \\
                                             & Hypersim-WB          & -      & -      & -     & -     & \textbf{2.19}          & \textbf{0.59}          & \textbf{4.65}          & \textbf{5.68}          & \textbf{3.16}          & \textbf{1.71}           & 5.19                   & 5.73                   \\\ \\\noalign{\vskip -1em} \cdashline{1-14} \\\noalign{\vskip -1em}
\multirow{2}{*}{FC$^4$~\cite{hu2017fc4}}                      & -                    & 4.02   & 1.72   & 7.34  & 8.62  & 4.53                   & 0.98                   & 10.30                   & 13.50                   & 3.84                   & 1.67                    & 6.62                   & 7.45                   \\
                                             & Hypersim-WB          & -      & -      & -     & -     & \textbf{1.95}          & \textbf{\uline{0.58}} & \textbf{4.30}           & \textbf{5.93}          & \textbf{3.37}          & \textbf{1.00}          & \textbf{6.60}           & \textbf{7.43}      \\ \bottomrule
\end{tabular}}
}

\end{table*}

%% file: sections/04_experimental_results.tex
\begin{figure*}[t]
\captionsetup[subfigure]{labelformat=empty}

\centering
    % Hypersim
    \begin{subfigure}{0.16\linewidth}
    \includegraphics[width=\linewidth]{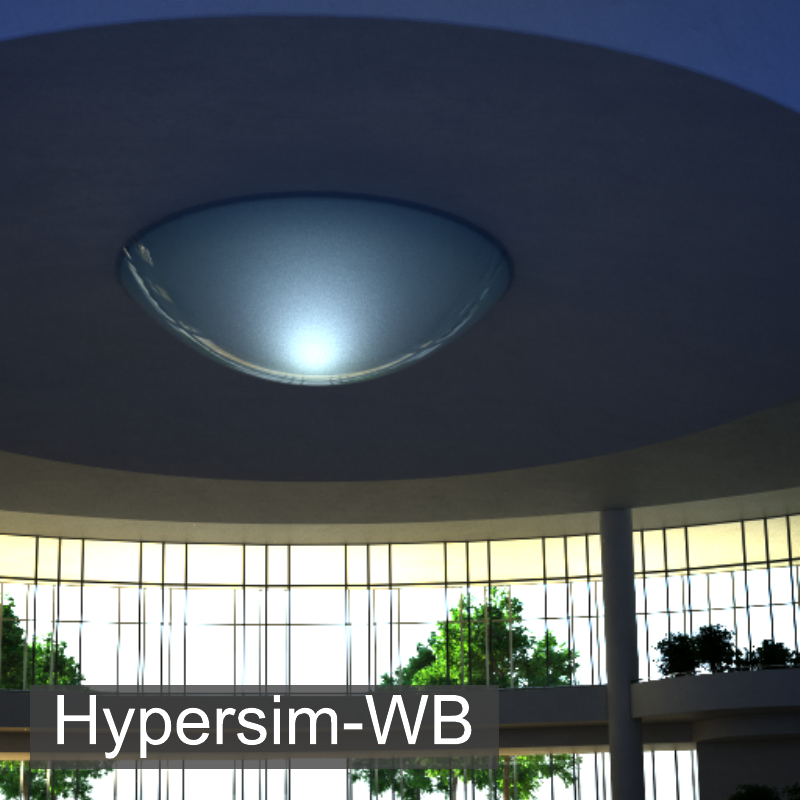} 
    \end{subfigure}
    \begin{subfigure}{0.16\linewidth}
    \includegraphics[width=\linewidth]{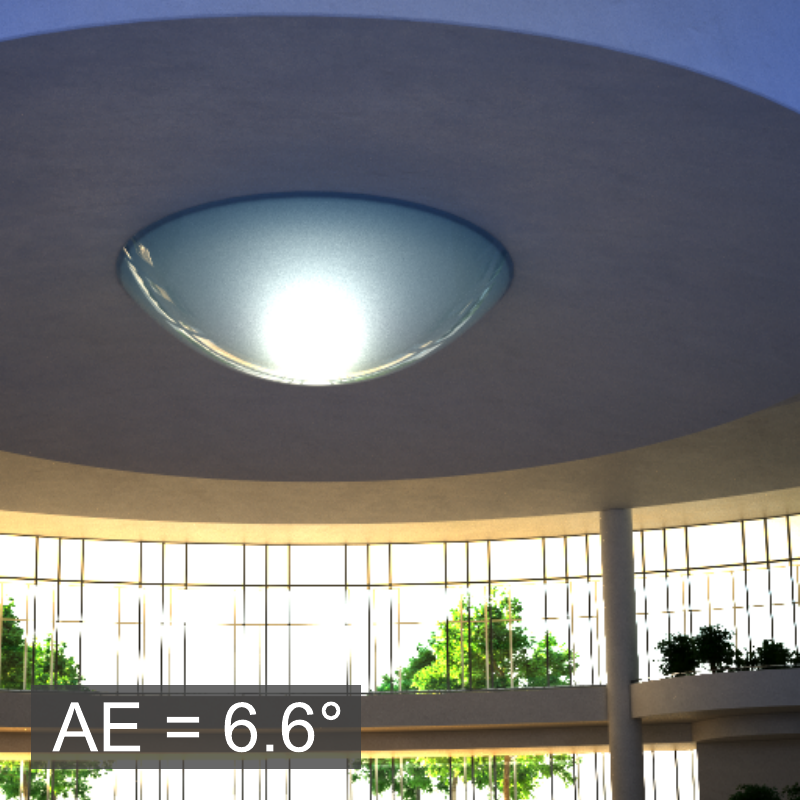} 
    \end{subfigure}
    \begin{subfigure}{0.16\linewidth}
    \includegraphics[width=\linewidth]{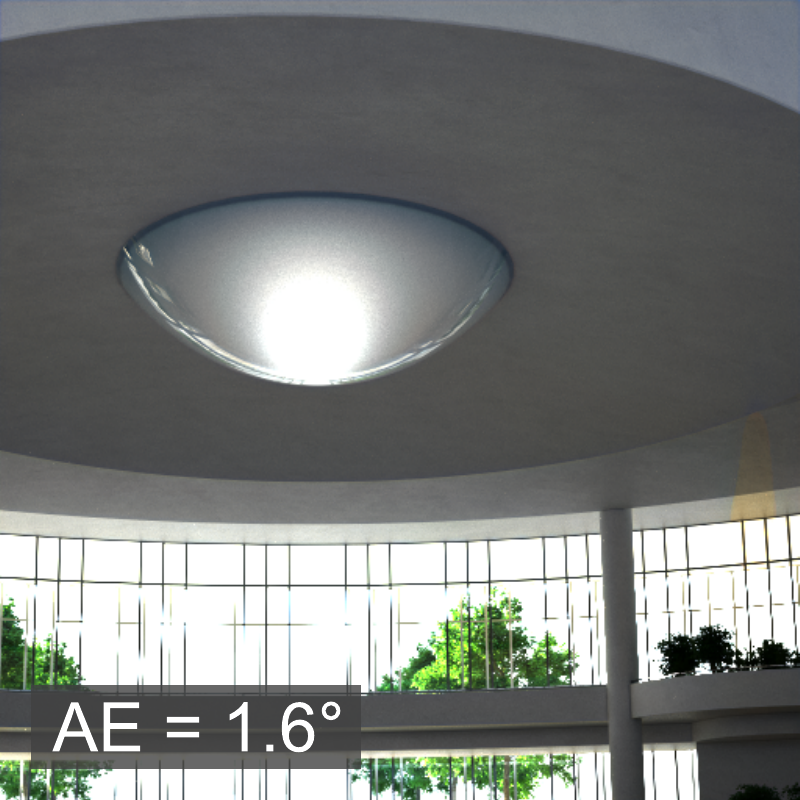} 
    \end{subfigure}
    \begin{subfigure}{0.16\linewidth}
    \includegraphics[width=\linewidth]{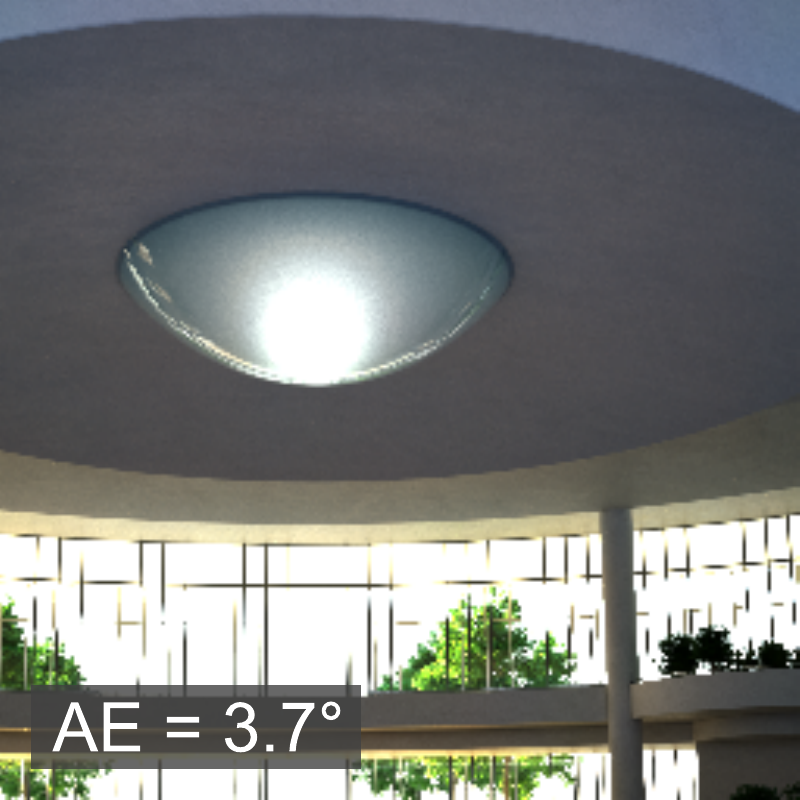} 
    \end{subfigure}
    \begin{subfigure}{0.16\linewidth}
    \includegraphics[width=\linewidth]{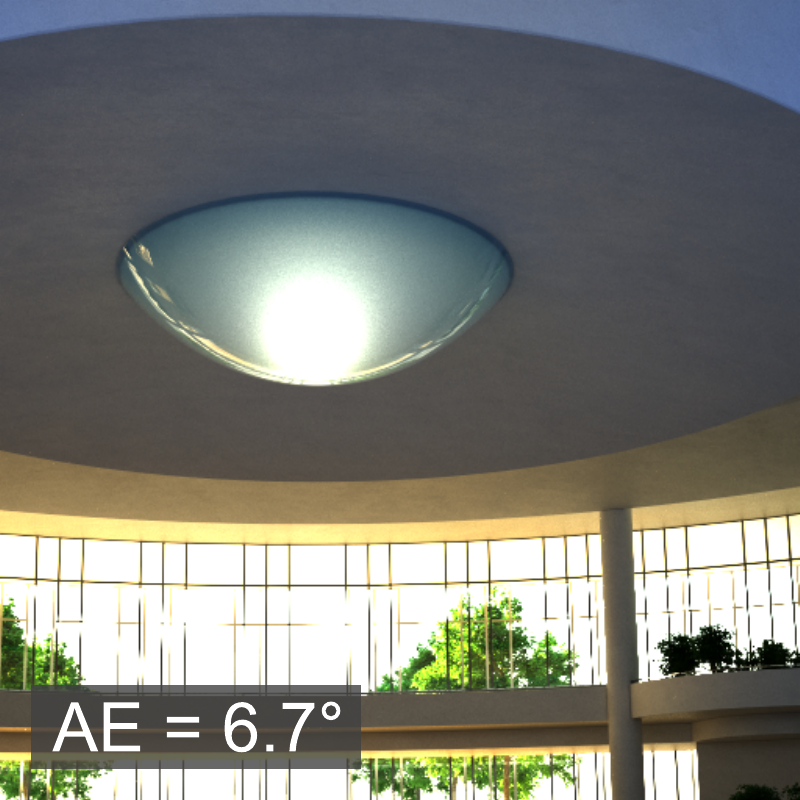} 
    \end{subfigure}
    \begin{subfigure}{0.16\linewidth}
    \includegraphics[width=\linewidth]{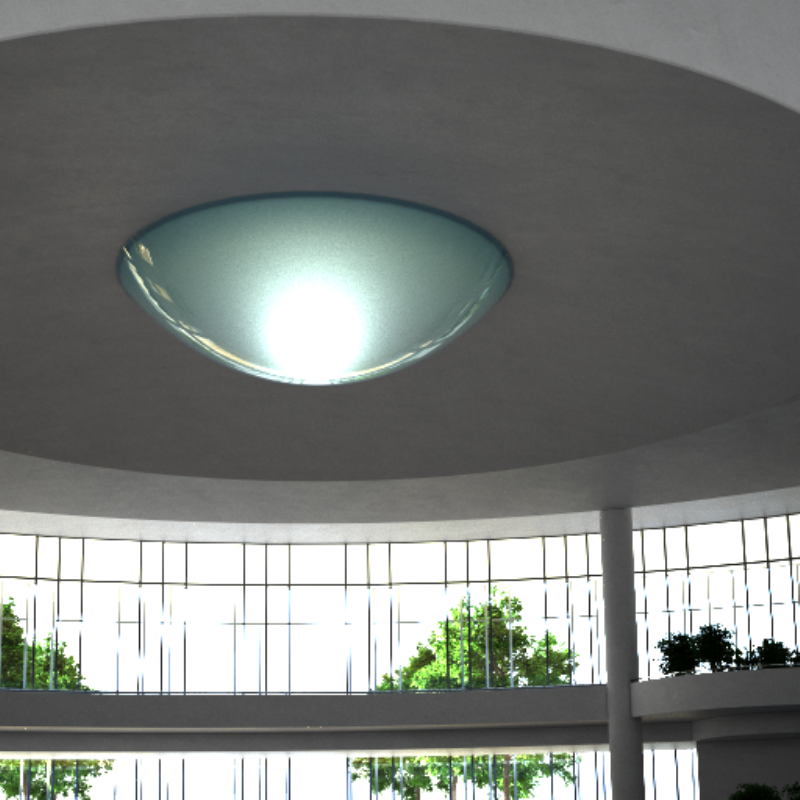}
    \end{subfigure}

    % LSMI
    \begin{subfigure}{0.16\linewidth}
    \includegraphics[width=\linewidth]{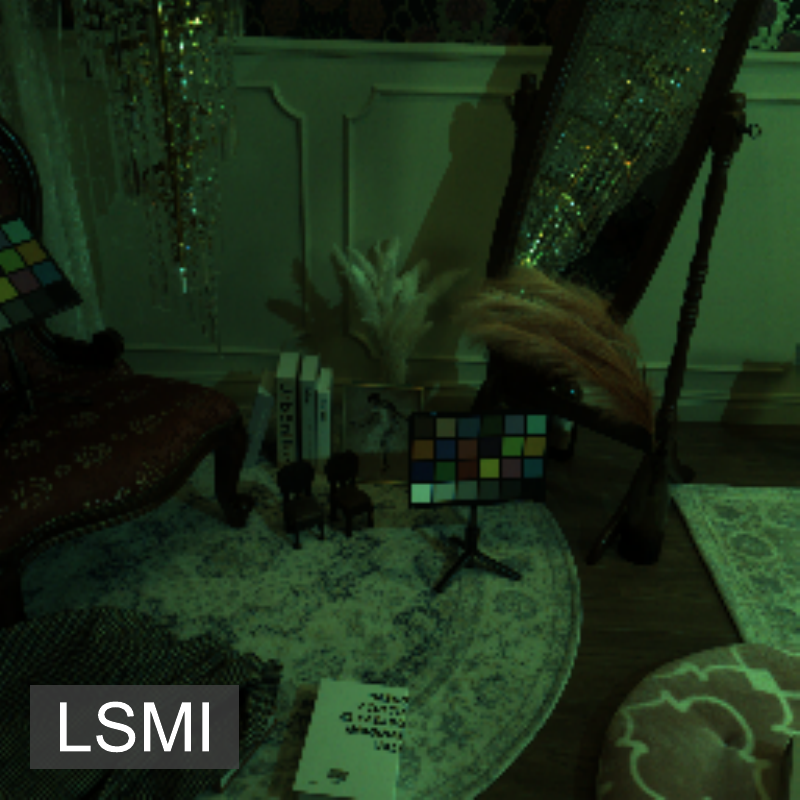} 
    \end{subfigure}
    \begin{subfigure}{0.16\linewidth}
    \includegraphics[width=\linewidth]{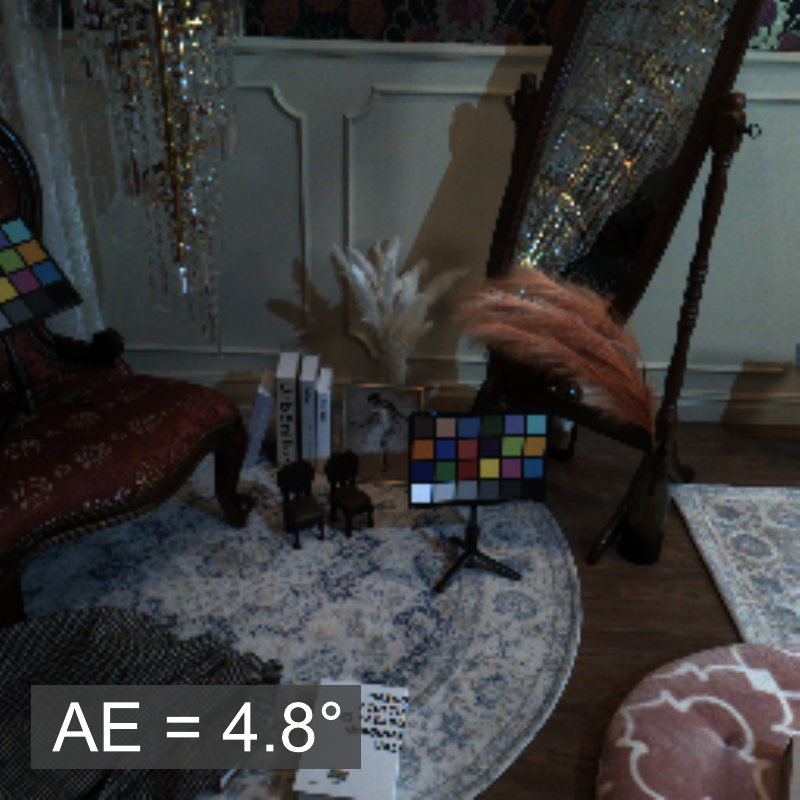} 
    \end{subfigure}
    \begin{subfigure}{0.16\linewidth}
    \includegraphics[width=\linewidth]{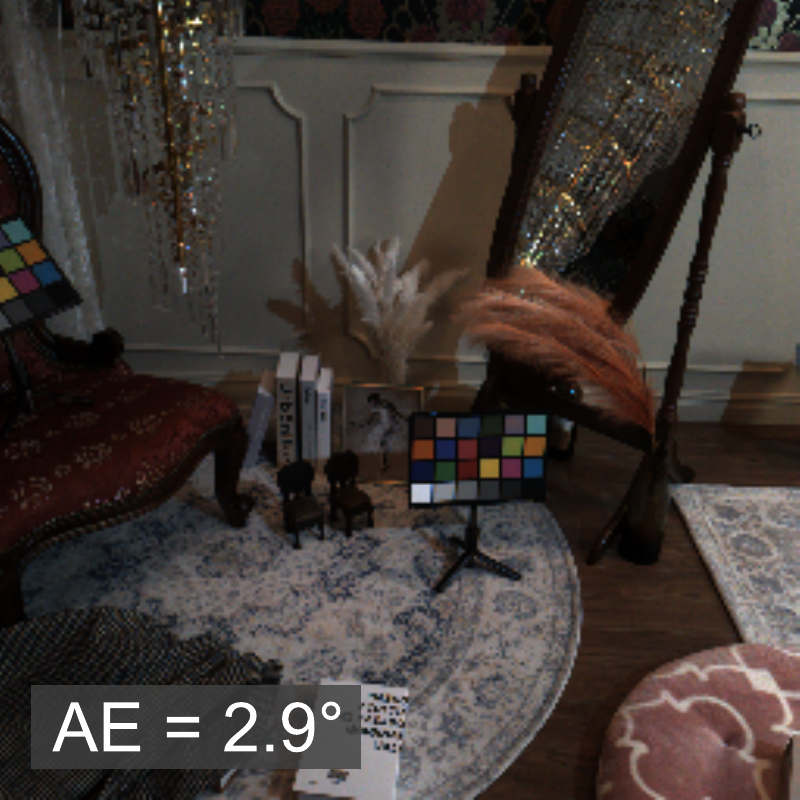} 
    \end{subfigure}
    \begin{subfigure}{0.16\linewidth}
    \includegraphics[width=\linewidth]{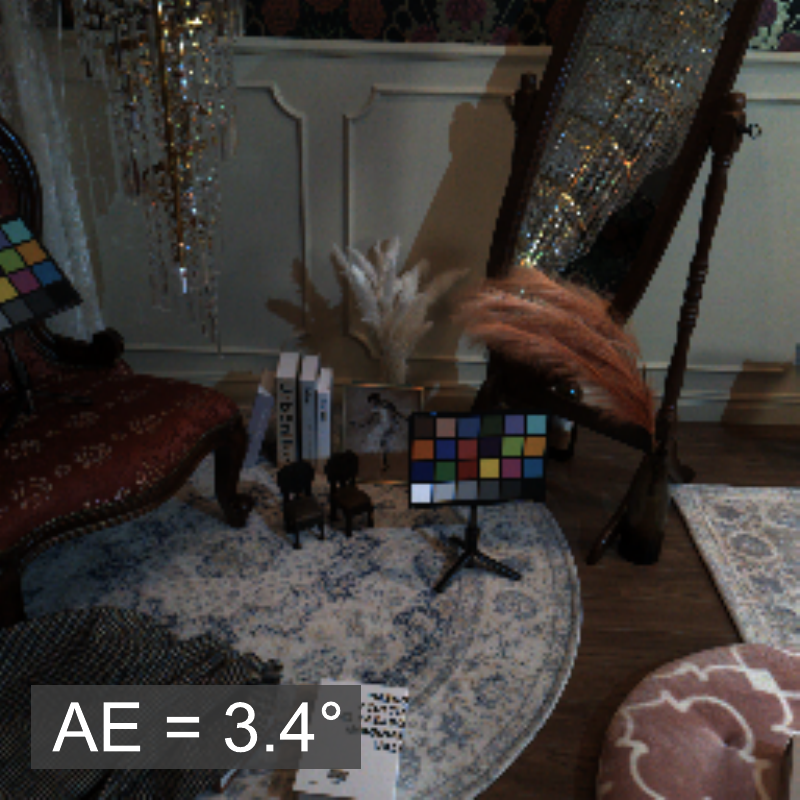} 
    \end{subfigure}
    \begin{subfigure}{0.16\linewidth}
    \includegraphics[width=\linewidth]{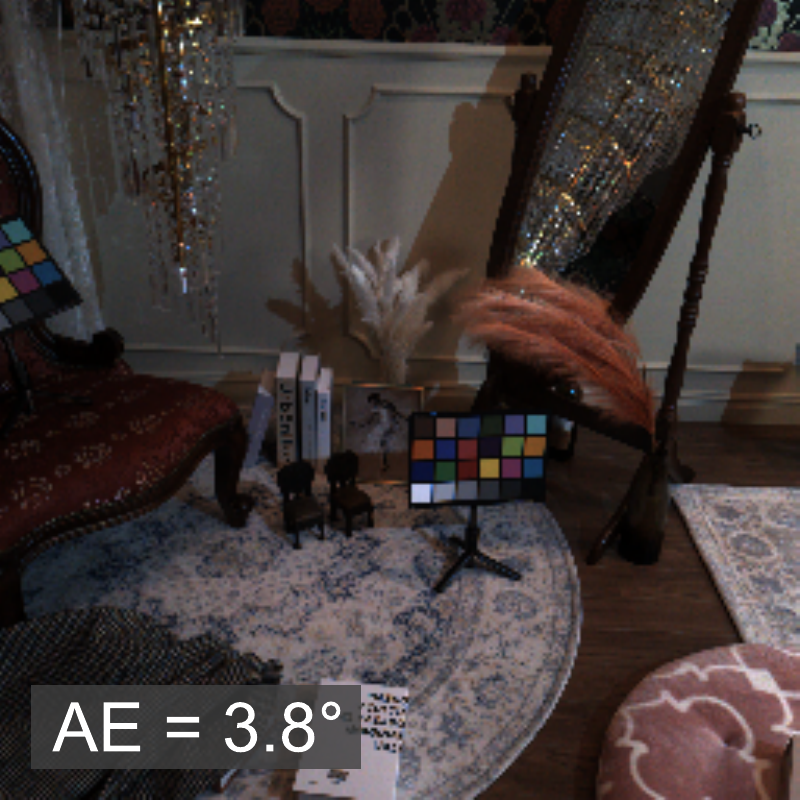} 
    \end{subfigure}
    \begin{subfigure}{0.16\linewidth}
    \includegraphics[width=\linewidth]{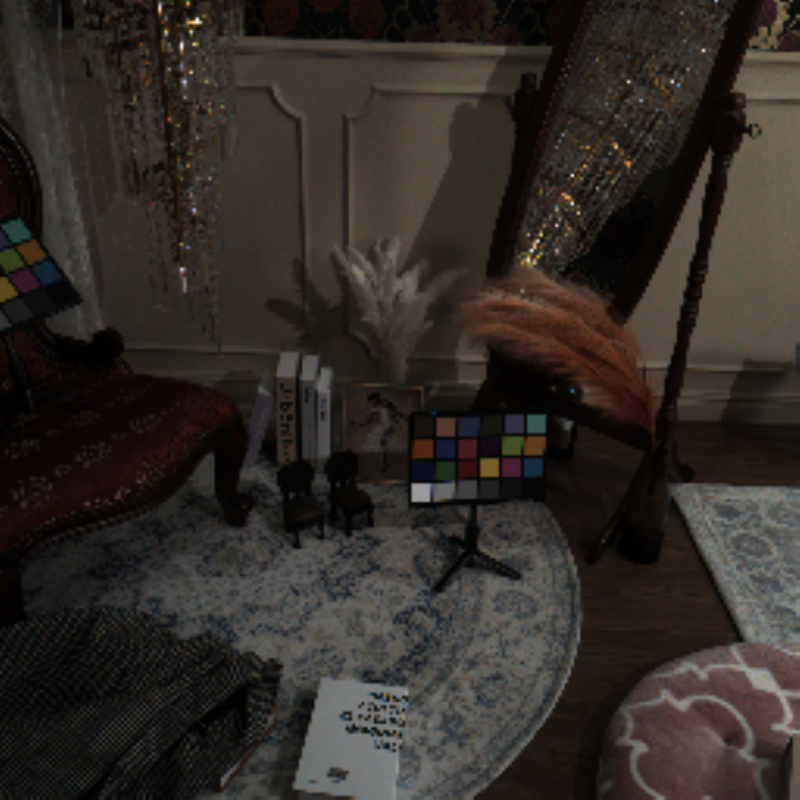}
    \end{subfigure}

    % MIMO
    \begin{subfigure}{0.16\linewidth}
    \includegraphics[width=\linewidth]{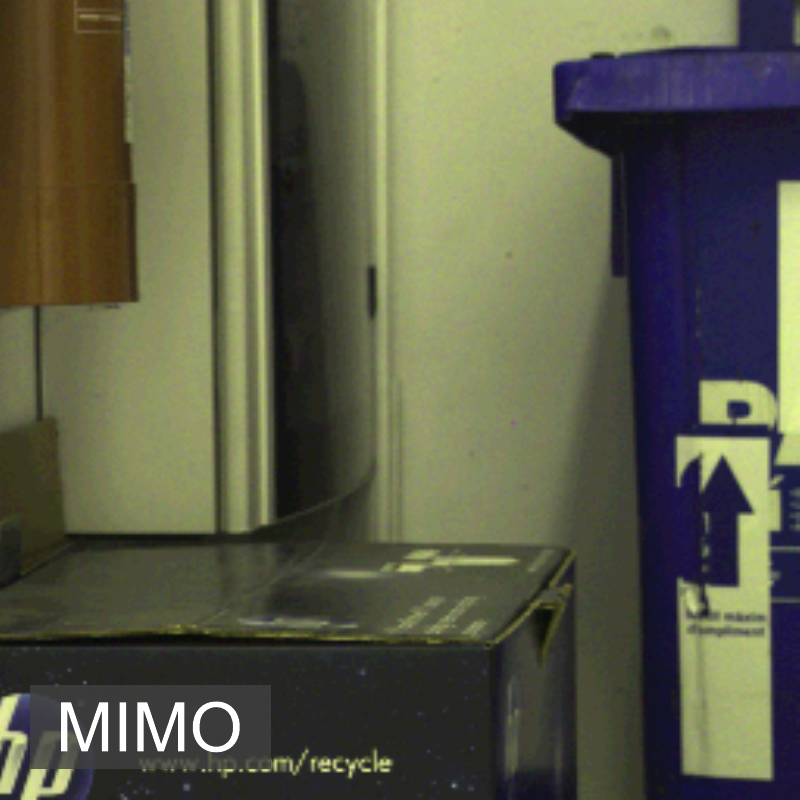} 
    \end{subfigure}
    \begin{subfigure}{0.16\linewidth}
    \includegraphics[width=\linewidth]{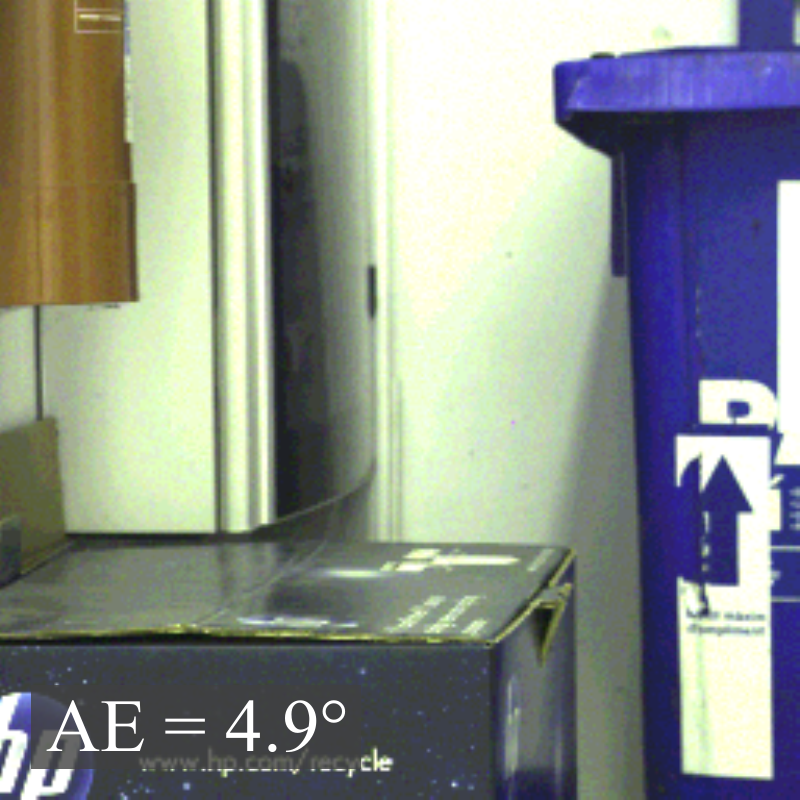} 
    \end{subfigure}
    \begin{subfigure}{0.16\linewidth}
    \includegraphics[width=\linewidth]{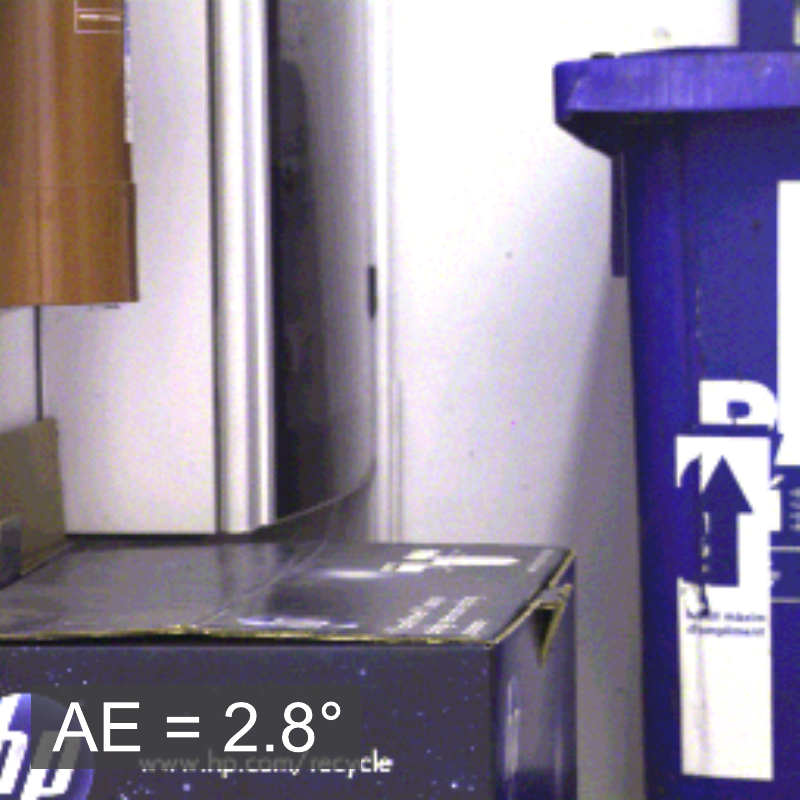} 
    \end{subfigure}
    \begin{subfigure}{0.16\linewidth}
    \includegraphics[width=\linewidth]{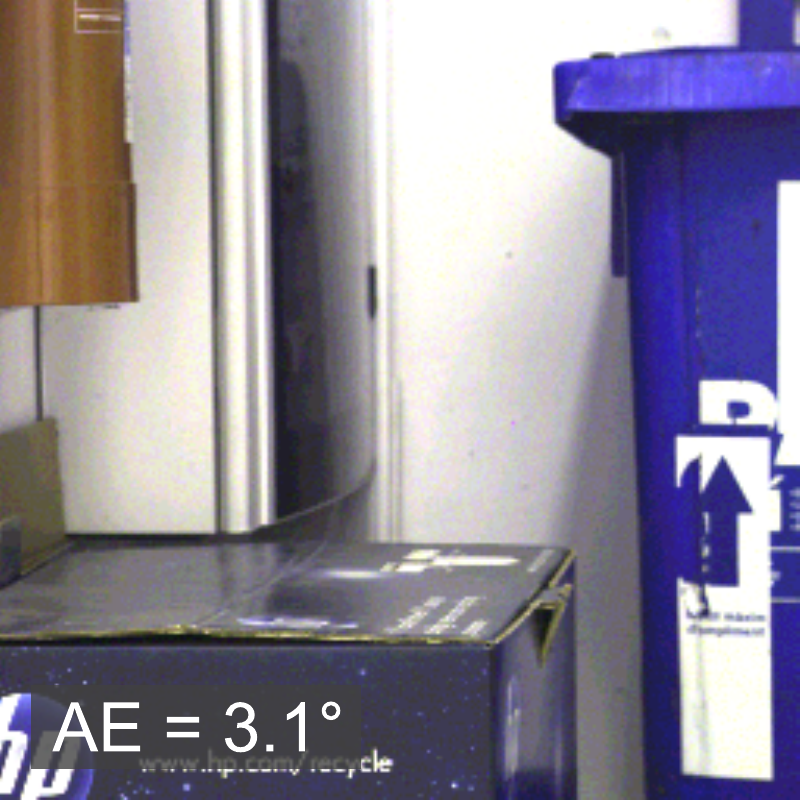} 
    \end{subfigure}
    \begin{subfigure}{0.16\linewidth}
    \includegraphics[width=\linewidth]{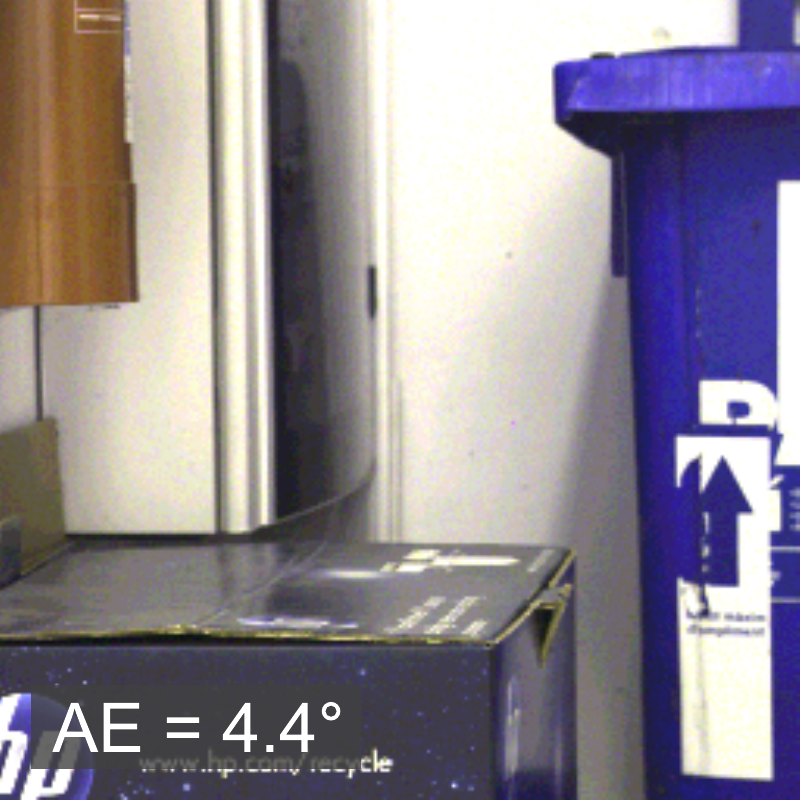} 
    \end{subfigure}
    \begin{subfigure}{0.16\linewidth}
    \includegraphics[width=\linewidth]{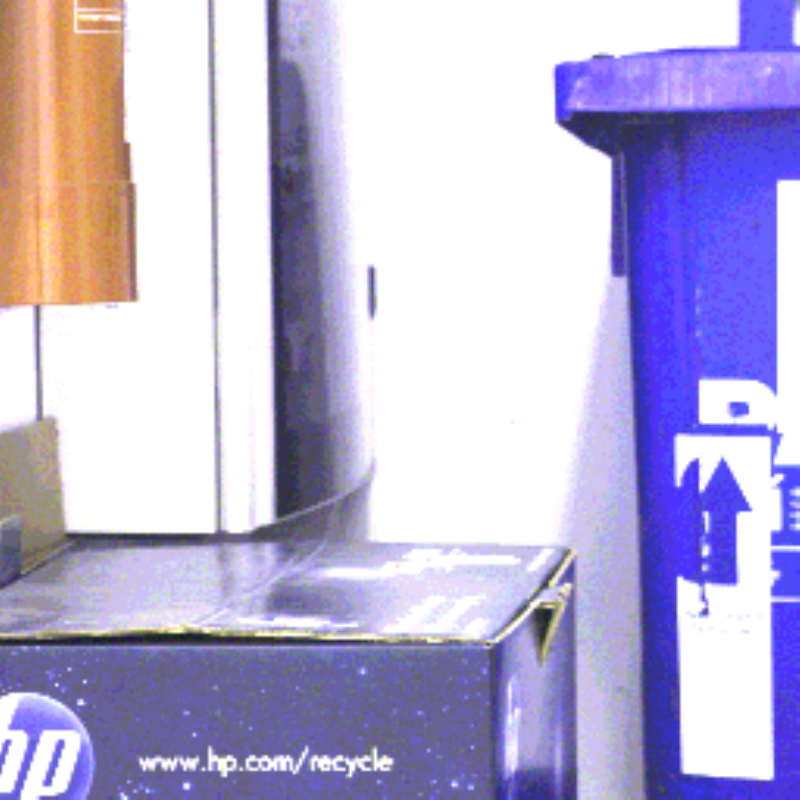} 
    \end{subfigure}

    % ColorChecker
    \begin{subfigure}{0.16\linewidth}
    \includegraphics[width=\linewidth]{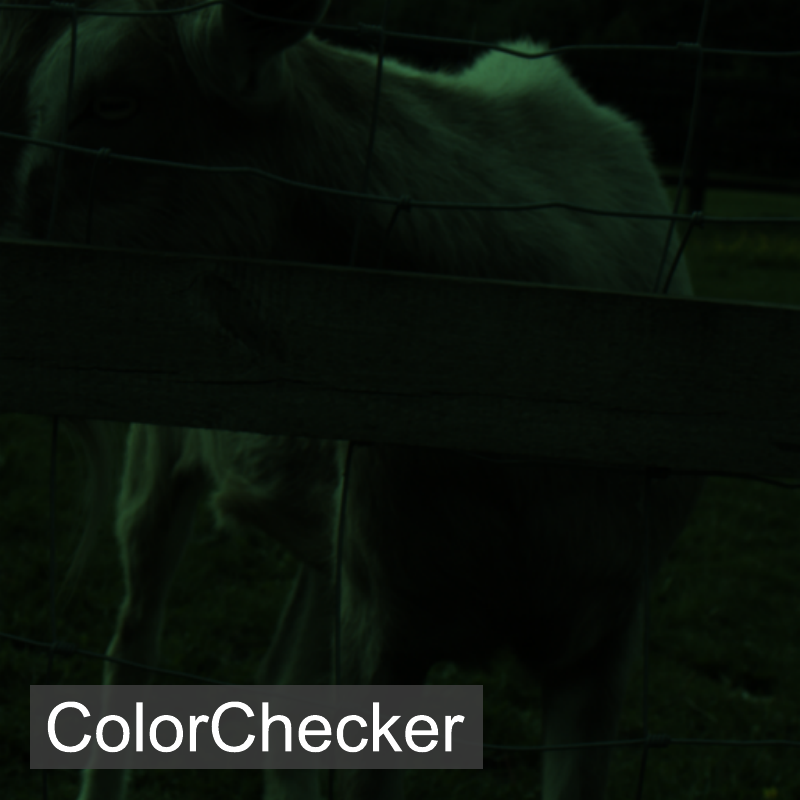} 
    \end{subfigure}
    \begin{subfigure}{0.16\linewidth}
    \includegraphics[width=\linewidth]{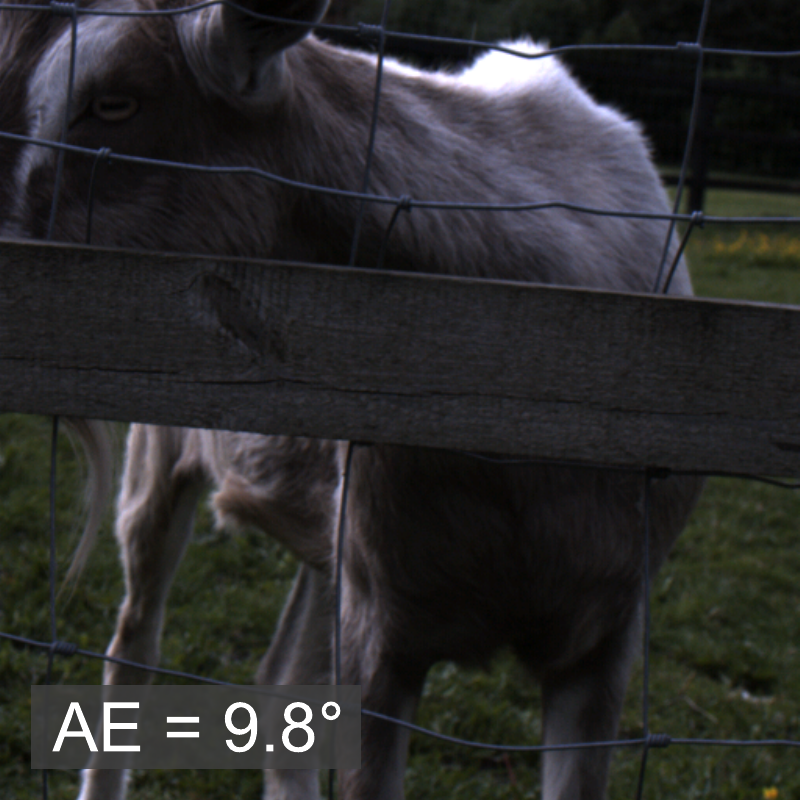} 
    \end{subfigure}
    \begin{subfigure}{0.16\linewidth}
    \includegraphics[width=\linewidth]{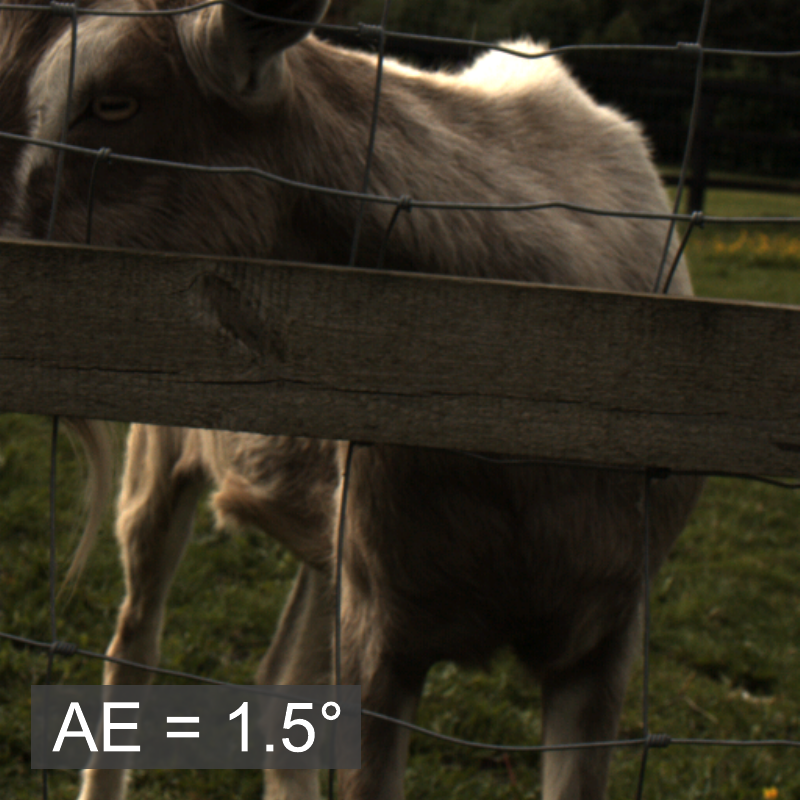} 
    \end{subfigure}
    \begin{subfigure}{0.16\linewidth}
    \includegraphics[width=\linewidth]{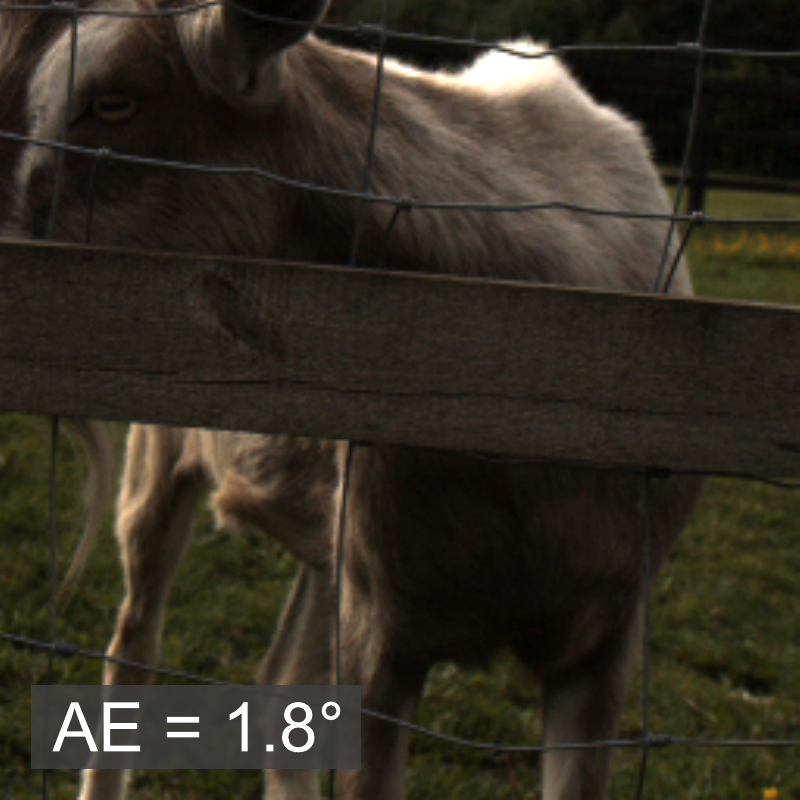} 
    \end{subfigure}
    \begin{subfigure}{0.16\linewidth}
    \includegraphics[width=\linewidth]{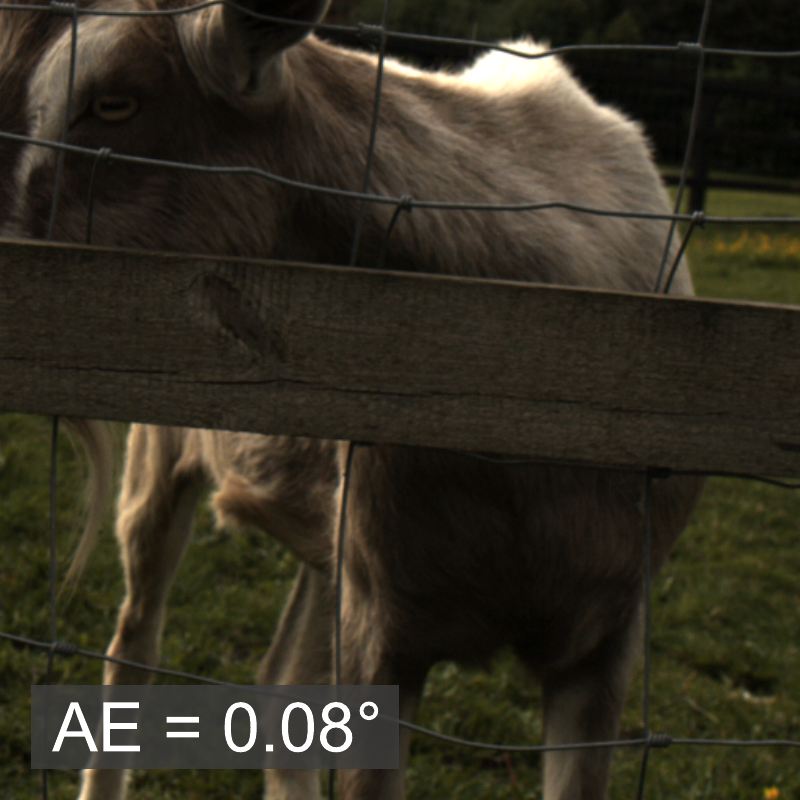} 
    \end{subfigure}
    \begin{subfigure}{0.16\linewidth}
    \includegraphics[width=\linewidth]{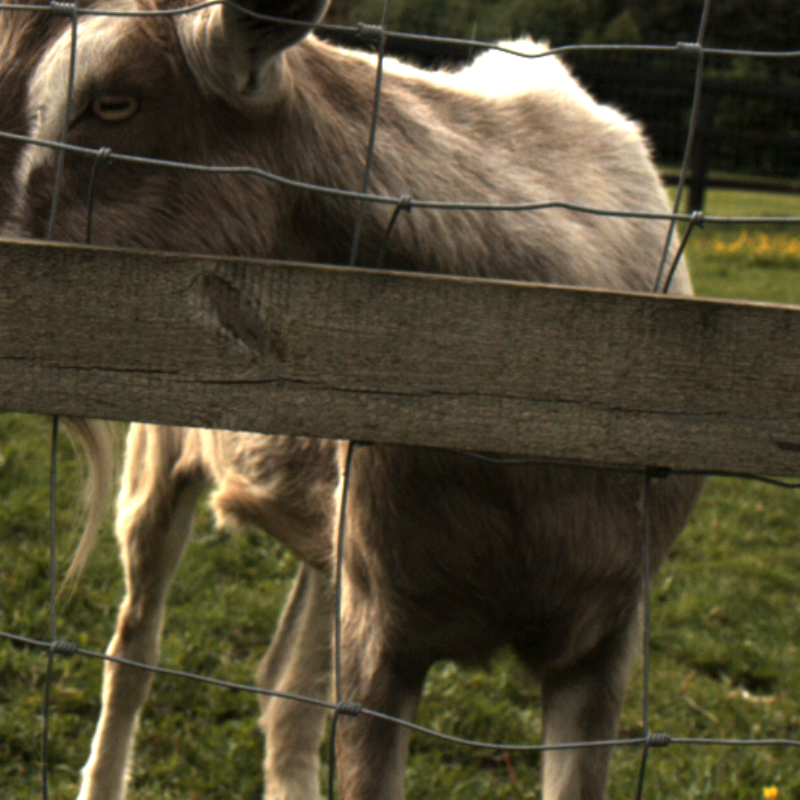} 
    \end{subfigure}

    % Captions row
    \vspace{-4mm}
    \subcaptionbox{\footnotesize Input}[0.16\linewidth]{}
    \subcaptionbox{\footnotesize GW~\cite{van2005edge}}[0.16\linewidth]{}
    \subcaptionbox{\footnotesize UNet~\cite{kim2021large}}[0.16\linewidth]{}
    \subcaptionbox{\footnotesize HDRNet~\cite{gharbi2017deep}}[0.16\linewidth]{}
    \subcaptionbox{\footnotesize FC$^4$~\cite{hu2017fc4}}[0.16\linewidth]{}
    \subcaptionbox{\footnotesize GT}[0.16\linewidth]{}
    \vspace{4mm}
\caption{Qualitative results of the methods on the considered datasets: 
Hypersim-WB, 
LSMI-Galaxy~\cite{kim2021large}, MIMO~\cite{gharbi2017deep}, and ColorChecker~\cite{hemrit2018rehabilitating}. The learning based methods from columns (3-5) are pretrained on Hypersim-WB and fine-tuned on the target dataset. The mean angular error for each image is also reported.}
\label{fig:main_qualitative}
\end{figure*}

\section{Experimental results}
\label{sec:experimental_results}
\subsection{Experimental Setting}
All experiments are conducted under a consistent setup to ensure fair comparisons between methods.
Our experimental protocol is designed to isolate the contribution of synthetic pre-training from architectural choices, focusing on data efficiency, robustness under limited supervision, and convergence behaviour rather than absolute peak performance. Models are first pre-trained on the Hypersim-WB dataset and subsequently fine-tuned on real-world datasets. We adopt the Adam optimizer~\cite{diederik2014adam} with an initial learning rate of $1\times10^{-4}$, scheduled via cosine annealing~\cite{loshchilov2016sgdr} with warm restarts every 10 epochs. During fine-tuning, early stopping is applied with a patience of 50 epochs.
For inference, test images are resized to $256\times256$ pixels, and model outputs are resized back to the original dimensions. Training is guided by the angular error loss, masking pixels where only the residual $\rho$ contributes to the radiance. Quantitative evaluation reports the angular error distribution, including mean, best-quartile (B-25), worst-quartile (W-25), and 95th percentile (95-P) values.

\subsection{Multi-Illuminant Estimation Experiments}
\label{sec:multi_illuminant_experiments}
For multi-illuminant estimation, we follow prior work~\cite{kim2021large} and evaluate two representative image-to-image architectures: UNet~\cite{ronneberger2015u} and HDRNet~\cite{gharbi2017deep}. Each model is trained to infer from RGB inputs the dense illuminant maps described in the Data Generation Pipeline section. For comparison, we also include the Gray World (GW)~\cite{van2005edge} algorithm and the FC$^4$ model~\cite{hu2017fc4}, representing statistical and learning-based single-illuminant methods, respectively. To evaluate these baselines, their global illuminant predictions are expanded to the size of the ground-truth dense map.
All learning-based models are initially trained on Hypersim-WB and then fine-tuned on the real-world LSMI~\cite{kim2021large} and MIMO~\cite{beigpour2013multi} datasets. To assess the effect of pre-training, we also train models from scratch on the target datasets. \Cref{tab:main_results} presents quantitative results, while \Cref{fig:main_qualitative} shows qualitative examples.
Results indicate that models pre-trained on Hypersim-WB consistently outperform those trained solely on real data. The improvement is particularly pronounced for smaller datasets like MIMO (78 images), highlighting the role of synthetic pre-training in enhancing generalization and convergence. Even for larger datasets such as LSMI, pre-trained models maintain superior performance compared to the trained-from-scratch counterparts.

\subsection{Dataset Reduction Experiment}
\label{sec:dataset_reduction_experiment}
\begin{figure}
    \centering
    \includegraphics[width=1\linewidth]{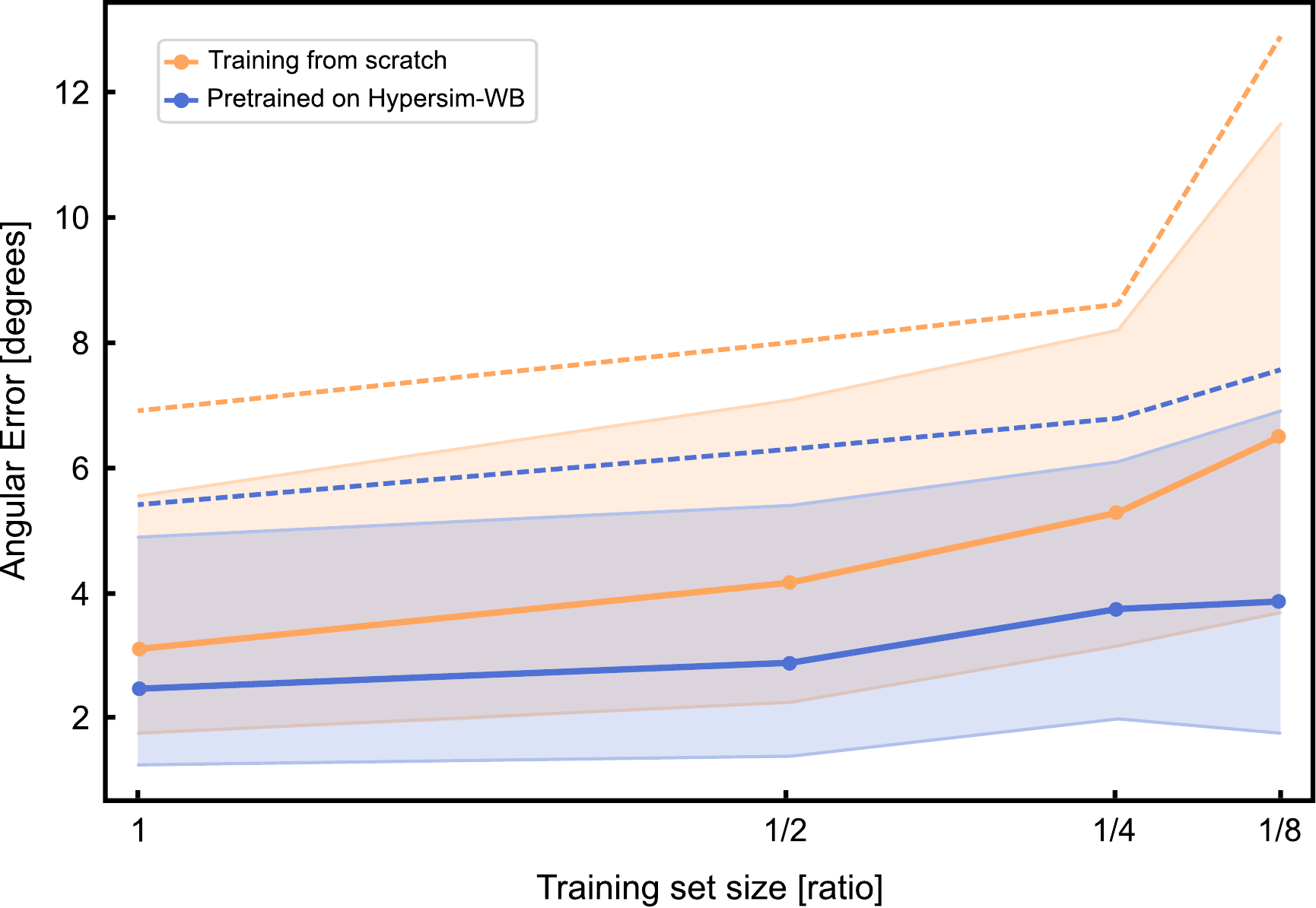}
    \vspace{1mm}
    \caption{Mean angular error for each considered training set size. 
    The upper and lower bound represent respectively the mean of the worst quartile (W-25) and best quartile (B-25) results.
    The dotted lines correspond to the mean 95th percentile.}
    \label{fig:fold_ae_plot}
\end{figure}

To better reflect realistic training conditions, we explicitly analyse the impact of synthetic pre-training under progressively reduced amounts of real data, a scenario that commonly arises in multi-illuminant estimation benchmarks.
We perform a dataset reduction study using the MIMO dataset. The training set is progressively reduced to 1/2, 1/4, and 1/8 of its original size, while the test set remains constant. Training follows a cross-fold validation scheme, and the UNet architecture is used, as it was the top-performing model in previous experiments.
We compare models pre-trained on Hypersim-WB with those trained from scratch. \Cref{fig:fold_ae_plot} shows the average angular error, best-quartile (B-25), worst-quartile (W-25), and 95th percentile values across different dataset cardinalities. As expected, performance degrades with decreasing dataset size, but fine-tuned models consistently outperform their from-scratch counterparts. Notably, pre-trained models exhibit a much slower degradation, demonstrating that synthetic pre-training provides robustness when real data are scarce.

\subsection{Single-Illuminant Estimation Experiment}
\label{sec:single_illuminant_experiment}
While Hypersim-WB is designed for multi-illuminant estimation with dense maps, it can be repurposed for single-illuminant tasks by averaging the ground-truth illuminant chromaticity over each image. We evaluate Gray World, FC$^4$~\cite{hu2017fc4}, and the UNet and HDRNet architectures (averaging their dense predictions to obtain a single RGB illuminant vector).
All models are pre-trained with the same hyperparameters described above and fine-tuned on the ColorChecker dataset~\cite{hemrit2018rehabilitating}. Quantitative results are reported in \Cref{tab:single_results}. While UNet and HDRNet perform suboptimally for single-illuminant estimation, fine-tuned versions still outperform models trained from scratch. For HDRNet, the from-scratch model achieves slightly better W-25 and 95th percentile values, but requires substantially longer training (359 epochs vs. 98 for the fine-tuned model), demonstrating that pre-training accelerates convergence while maintaining comparable performance.
For FC$^4$, fine-tuning on Hypersim-WB consistently lowers the angular error compared to training from scratch, confirming that synthetic pre-training improves performance even in single-illuminant estimation.

%% file: sections/05_conclusion.tex
\section{Conclusions}
\label{sec:conclusion}

In this work, we address data scarcity in illuminant estimation by proposing a general methodology for deriving dense illuminant supervision from physically based 3D-rendered scenes. By exploiting physically grounded illumination components, the proposed pipeline enables the extraction of pixel-wise illuminant chromaticity maps for supervised single- and multi-illuminant estimation. As a concrete instantiation, we apply it to the Hypersim 3D scene collection to generate large-scale synthetic training data.
Extensive experiments with state-of-the-art architectures, including UNet, HDRNet, and FC$^4$, demonstrate that pre-training on the proposed synthetic dataset consistently improves performance when models are fine-tuned on real benchmarks. Gains are observed across datasets and tasks, with particularly strong improvements in multi-illuminant estimation and under limited real-data conditions. Moreover, synthetic pre-training accelerates convergence and enhances model robustness.
These results highlight that physically based synthetic data offer an effective solution to the limitations imposed by scarce real-world annotations in illuminant estimation. Future work includes applying the pipeline to datasets featuring outdoor and more complex lighting conditions, as well as investigating domain adaptation to better transfer learned illuminant representations to real-world images.

\input{tabs/quantitative_single}

%% file: tabs/quantitative_single.tex
\begin{table}
\caption{Quantitative results, in terms of angular error, of the considered methods for the task of single-illuminant estimation on the ColorChecker dataset~\cite{hemrit2018rehabilitating}. For each learning-based method we compare a version trained from-scratch and one that is fine-tuned after pre-training on the Hypersim-WB dataset. We highlight in bold the per-method best and underline the per-dataset best.\vspace{2mm}}
\label{tab:single_results}
\resizebox{\linewidth}{!}{
{\sffamily
\begin{tabular}{lcllll}
\toprule
                                             &  & \multicolumn{4}{c}{ColorChecker~\cite{hemrit2018rehabilitating}}                                                                          \\ \cmidrule(lr){3-6} \\\\\noalign{\vskip -2.5em}
Method                                       & Pre-training Data    & \multicolumn{1}{c}{Mean} & \multicolumn{1}{c}{B-25} & \multicolumn{1}{c}{W-25} & \multicolumn{1}{c}{95-P} \\ \hline \noalign{\vskip 0.25em}
GW~\cite{van2005edge}     & -                    & 4.80                      & 1.72                     & 7.11                     & 13.20                     \\\ \\\noalign{\vskip -1em} \cdashline{1-6} \\\noalign{\vskip -1em}
\multirow{2}{*}{UNet~\cite{kim2021large}}                        & -                    & 3.14                     & 0.69                    & \textbf{6.72}            & 8.84                     \\
                                             & Hypersim-WB          & \textbf{2.94}            & {\uline{\textbf{0.63}}}      & 6.76                     & \textbf{7.55}            \\\ \\\noalign{\vskip -1em} \cdashline{1-6} \\\noalign{\vskip -1em}
\multirow{2}{*}{HDRNet~\cite{gharbi2017deep}}                      & -                    & 3.53                     & 0.90                    & \textbf{7.20}             & \textbf{9.25}            \\
                                             & Hypersim-WB          & \textbf{3.40}             & \textbf{0.82}           & 7.46                     & 9.50                      \\\ \\\noalign{\vskip -1em} \cdashline{1-6} \\\noalign{\vskip -1em}
\multirow{2}{*}{FC$^4$~\cite{hu2017fc4}}                      & -                    & 4.02                     & 1.36                     & 8.29                     & 9.76                     \\
                                             & Hypersim-WB          & {\uline{\textbf{2.88}}}      & \textbf{0.74}           & {\uline{\textbf{6.19}}}      & {\uline{\textbf{7.09}}}    \\ \bottomrule 
\end{tabular}}
}

\end{table}